\documentclass[conference]{IEEEtran}
\IEEEoverridecommandlockouts

\usepackage{cite}
\usepackage{amsmath,amssymb,amsfonts}
\usepackage{balance}
\usepackage{algorithmic}
\usepackage{algorithm}
\usepackage{graphicx}
\usepackage{textcomp}
\usepackage{xcolor}
\usepackage{url}
\usepackage{booktabs}
\usepackage{multirow}
\usepackage{placeins}
\usepackage{enumitem}
\usepackage[hidelinks]{hyperref}

\graphicspath{{figures/}}

\newcommand{\pal}{PAL-Bench}
\newcommand{\paltrace}{PAL-TRACE}
\newcommand{\nllm}{\textsc{nLLM}}

\def\BibTeX{{\rm B\kern-.05em{\sc i\kern-.025em b}\kern-.08em
    T\kern-.1667em\lower.7ex\hbox{E}\kern-.125emX}}

\begin{document}

\title{[Experiment, Analysis, and Benchmark] PAL-Bench: Evidence-Grounded Profile Reconstruction from Longitudinal Personal Albums}

\author{
\IEEEauthorblockN{
Qiwei Yan\textsuperscript{1,2,$\dagger$},
Zhiqiang Yuan\textsuperscript{1,$\dagger$ $\S$},
Zexi Jia\textsuperscript{1},
Nanxing Hu\textsuperscript{1},
Kailin Lyu\textsuperscript{3},
Jie Zhou\textsuperscript{1},
Jinchao Zhang\textsuperscript{1*}
}
\IEEEauthorblockA{
\textsuperscript{1}WeChat AI, Tencent, Beijing, China\\
\textsuperscript{2}University of Chinese Academy of Sciences, Beijing, China\\
\textsuperscript{3}Institute of Automation, Chinese Academy of Sciences, Beijing, China\\
yanqiwei22@mails.ucas.edu.cn, yuanzhiqiang19@mails.ucas.ac.cn, dayerzhang@tencent.com
}
}

\maketitle

\begingroup
\renewcommand{\thefootnote}{\fnsymbol{footnote}}
\footnotetext[2]{Equal contribution. $^{\S}$ Tech Lead. * Corresponding authors. This work was done during Qiwei Yan's internship at WeChat AI, Tencent Inc. under the guidance of Zhiqiang Yuan.}
\endgroup

\begin{abstract}
Longitudinal personal albums are weak-schema multimodal databases: their records expose noisy perceptual fields, while their most important facts require joins across faces, text, timestamps, locations, and repeated events. Existing visual, video, document, and lifelog benchmarks test sub-problems, but not album-scale profile reconstruction with social identity binding and evidence citation. Benchmarking this task is difficult because the complete ground truth needed for evaluation---owner profiles, social graphs, face-name maps, and evidence provenance---is exactly the private state that real albums cannot safely release. We introduce \pal{}, a controlled benchmark for evidence-grounded reconstruction under a public-record contract. Its Evidence Compiler builds latent private worlds, programs target-level evidence paths, renders album pixels, re-measures them through perception pipelines, and exports audited public/private views. Agents receive only perception-derived public records; targets, identifier maps, and evidence paths remain hidden. \pal{} contains 50 synthetic users, 36,659 public photo records, and 2,799 targets over owner facts, identities, and relations. A 10-participant privacy-preserving audit establishes construct coverage: the evidence structures \pal{} controls occur in real private albums, even though equivalent real releases are privacy-prohibitive. Across seven system classes and two compute-matched diagnostics, a seven-metric protocol reveals a gap between plausible profile summarization and faithful social reconstruction: systems recover some owner facts but struggle with recurring identities and evidence citation. \paltrace{}, a reference framework that freezes identity bindings before owner-fact mining, performs best but leaves hard identity resolution far from solved. \pal{} provides a controlled testbed for perceptual entity resolution, multimodal data integration, temporal evidence aggregation, and provenance-aware structured prediction. Code and data are available at \textcolor{blue}{\href{https://github.com/icde2027-palbench/pal-bench}{[link]}}.
\end{abstract}


\section{Introduction}
\label{sec:introduction}

\begin{figure}[t]
\centering
\includegraphics[width=\linewidth]{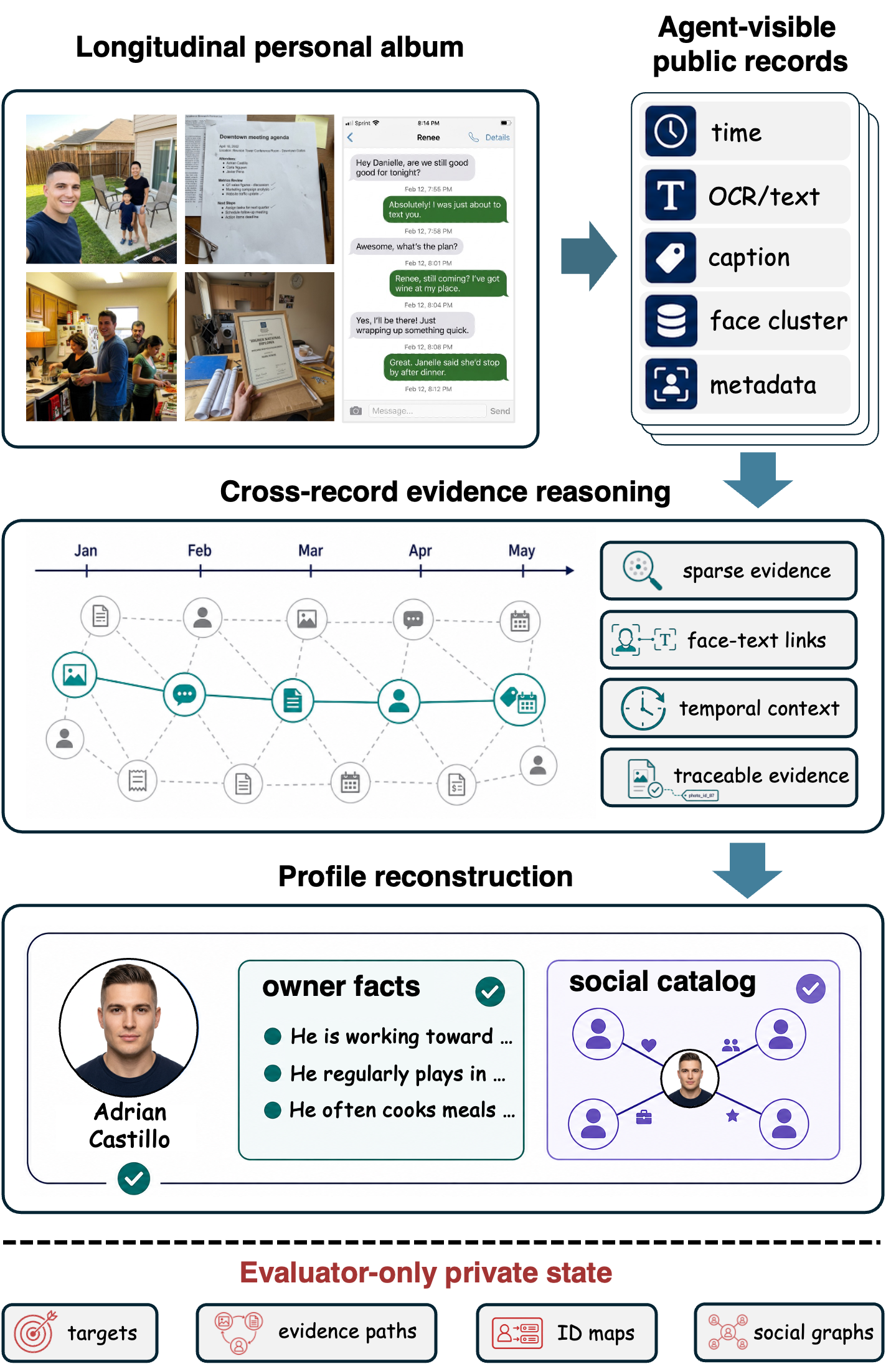}
\caption{\pal{} task contract. The agent observes public album records and perception-derived fields, performs cross-record evidence reasoning, and reconstructs an evidence-cited owner profile and social catalog; targets, evidence paths, ID maps, and social graphs remain evaluator-only private state. All shown photos, persons, and profile details are synthetic; no real personal album content is displayed.}
\vspace{-15px}
\label{fig:teaser}
\end{figure}

A personal album is a longitudinal, weak-schema multimodal database. Receipts, badges, chat screenshots, travel scenes, family gatherings, commute routes, recurring faces, calendars, meals, devices, and locations are not clean relational tuples, but they jointly encode facts about an owner: occupation, city, hobbies, routines, relationship structure, and the identities of recurring people. The central data-management operation is therefore not recognizing one image in isolation, but constructing reliable joins across noisy perceptual records---faces, OCR strings, captions, timestamps, locations, and event co-occurrence patterns---and turning those joins into structured claims with public evidence.

This task is straightforward to describe but difficult to benchmark. Complete evaluation requires private state---owner profiles, social graphs, face-name mappings, and per-target evidence provenance---that real personal albums cannot ethically expose. At the same time, naive synthetic data risks rewarding hidden generator intent rather than observable evidence. A benchmark for this setting must therefore be \emph{complete} enough to evaluate structured reconstruction and \emph{auditable} enough to show that systems solve the task from public observations rather than leaked private state.

Existing benchmarks address important sub-problems but not the integrated task. Visual question answering (VQA) and scene-graph tasks~\cite{antol2015vqa,johnson2017clevr,hudson2019gqa,marino2019okvqa} focus on answering prompted questions over one image or a small visual context. Optical character recognition (OCR) and document-image benchmarks~\cite{singh2019textvqa,mathew2021docvqa,biten2019stvqa} test text extraction and local visual-textual reasoning. Long-video and egocentric benchmarks~\cite{lei2018tvqa,grauman2022ego4d,damen2018epickitchens,wu2024longvideobench,fu2025videomme} introduce temporal context but evaluate event comprehension rather than open-world profile reconstruction; multi-image benchmarks~\cite{song2024milebench,wang2024muirbench} stress cross-frame reasoning at much smaller scale (typically tens of images per query). Lifelog benchmarks~\cite{gurrin2016ntcir,jiang2017memexqa,doherty2024memoriqa,tonellotto2023timelineqa} are closest in domain, yet they do not provide complete social identity ground truth, programmed evidence paths, public/private separation, or evidence-cited structured prediction at album scale.
To our knowledge, no existing benchmark evaluates this full contract: open-world reconstruction of an owner profile and social catalog from album-scale multimodal records, with hidden social ground truth, public evidence citations, and audited public/private separation.

We introduce \pal{}, a benchmark and empirical study for \emph{longitudinal profile reconstruction} from personal albums (Fig.~\ref{fig:teaser}). The system observes only public album records (timestamps, coarse metadata, captions, OCR strings, text entities, remapped face clusters) and must output owner facts and social-person rows with confidence and cited public evidence; evaluation uses hidden targets, private-to-public identifier maps, and audited evidence paths.

The central construction idea is \emph{evidence compilation}: \pal{} first builds a latent private world (owner profile, social graph, longitudinal timeline), programs evidence paths that specify how each target should become inferable, renders photos around those paths, measures the rendered images through perception models, and exports separated public/private views. Synthetic construction is not a convenience here but the enabling condition for a releasable benchmark contract: the evaluator can hold complete private ground truth, while the system sees only public observations. The renderer is used only as an evidence realizer, not as an oracle: Gemini 3.1 Flash Image (Nano Banana 2; API model ID: \texttt{gemini-3.1-flash-image})~\cite{google2026geminiimage} renders album pixels, while all agent-visible fields are rebuilt by separate perception measurements. Every target has a planned evidential route; every public field is perception-derived or release-safe; every evaluation reference resolves to a public record.

\pal{} is also an empirical benchmark, not only a dataset release. We evaluate seven system classes (deterministic heuristic, text-only large language model (LLM), multimodal retrieval-augmented generation (RAG), long-context multimodal LLM, generic tool-use, an adapted lifelog/profile baseline, and \paltrace{}) plus two compute-matched diagnostics that hold call budget fixed against \paltrace{}, all under a unified contract. \paltrace{}, our reference framework, instantiates a database-style principle: identity bindings should be stabilized as persistent entity state before type-open owner-fact mining proceeds. Long-context and RAG systems retrieve plausible facts but struggle to bind identities and cite faithful evidence; matched-call diagnostics show that call budget alone is insufficient when identity state is repeatedly regenerated in fresh prompt contexts. \paltrace{} improves owner fact recovery and evidence faithfulness, but hard person identity resolution remains far from solved. \pal{} thus exposes a set of data-engineering bottlenecks: sparse retrieval, perceptual entity resolution, temporal alignment, and provenance-aware composition.

\smallskip\noindent\textbf{Contributions.} (i) We formalize longitudinal profile reconstruction from personal albums as open-world structured prediction over public multimodal records, requiring fact recovery, face-name resolution, relation inference, confidence, and evidence citation. (ii) We introduce an \emph{Evidence Compiler} that turns latent private worlds into rendered, perception-measured public observations with audited public/private separation. (iii) We release a 50-user benchmark with 36,659 public records, 2,799 evaluation targets covering 840 owner-fact atoms and 653 social persons (each owner-fact target backed by a programmed cross-modal evidence path; identity targets jointly score name, relation, and category for the same person), accompanied by a 10-participant construct-coverage audit showing that the controlled evidence structures occur in real private albums while privacy barriers prevent equivalent real releases. (iv) We provide a seven-metric protocol and a multi-system experiment, analysis, and benchmark (EAB) study; \paltrace{}, a reference framework built around system-level state separation between identity bindings and owner facts, shows how database-style identity discipline can turn multi-call compute into capability gains, while identity grounding and evidence faithfulness remain primary bottlenecks.

\section{Task and Benchmark Contract}
\label{sec:task}

\subsection{Album Observations}
\label{subsec:album_model}

For each user $u$, \pal{} defines a public album
$A_u = \langle p_1, p_2, \ldots, p_n \rangle$ ordered by timestamp. Each record
$p_i = (\textit{id}_i, \tau_i, \mathbf{m}_i, \mathbf{v}_i)$ contains:
\begin{itemize}[leftmargin=*,nosep]
\item an opaque public photo identifier and timestamp $\tau_i$;
\item release-safe metadata $\mathbf{m}_i$ such as coarse location and device;
\item perception-derived content $\mathbf{v}_i$, i.e., outputs produced by face detection/matching, image captioning, OCR, and entity extraction applied to the rendered image, including caption, OCR strings, text entities, and public face identifiers.
\end{itemize}
Agents never receive private generation prompts, event roles, key-photo labels, target lists, private person identifiers, or private evidence paths. These remain in a hidden evaluator-only state that also stores original manifest/render IDs, latent owner timelines, and support sets.


\subsection{Targets and Outputs}
\label{subsec:targets}

The hidden evaluation state defines 840 owner fact atoms and three target families over the same 653 social persons---name, relation, and category---for 2,799 targets total. Owner facts are decomposed into atoms so that compound statements are scored precisely; relation/category credit is conditioned on correct identity binding (\S\ref{subsec:metrics}).

Given $A_u$, a system outputs
\begin{equation}
\hat{Y}_u = (\hat{O}_u, \hat{S}_u, \hat{C}_u),
\end{equation}
where $\hat{O}_u$ is a set of predicted owner facts, $\hat{S}_u$ is a set of predicted social-person rows, and $\hat{C}_u$ attaches confidence scores and cited public evidence to each claim. A social row must include a public face cluster when the prediction is visually grounded; relation and category credit is conditioned on correct identity binding.

The task differs from standard QA in two ways: there is no per-target question (the agent must discover what is recoverable), and the output is a structured profile with evidence rather than a free-form answer. This places \pal{} closer to data integration and knowledge construction~\cite{halevy2006data,dong2014knowledge} than to single-question visual reasoning. Formally, each hidden target $t \in \mathcal{T}_u$ has a private support set $E_t \subseteq A_u$ and a reasoning path $\rho_t$ over modalities (text, face, time, metadata, scene context). A valid prediction must (i) state a claim semantically equivalent to $t$, (ii) bind any required entity to the correct public identifier, and (iii) cite public records whose measured content supports the claim. This separation isolates fact knowledge, identity binding, and evidence support as three failure modes.

\subsection{Public/Private Separation}
\label{subsec:separation}

The benchmark contract separates the public agent input from the private evaluator state. The public album view contains only observable fields. The private evaluator view contains owner profiles, social graphs, public-to-private identifier maps, target definitions, planned reasoning paths, and audit metadata. Four invariants are enforced:
\begin{enumerate}[leftmargin=*,nosep]
\item public content derives only from perception outputs or release-safe metadata;
\item private identifiers, event IDs, roles, and key-photo labels never appear publicly;
\item face identifiers are remapped before release;
\item every target and evidence reference resolves to an existing public photo.
\end{enumerate}
All 50 released users pass this dual-view audit. We do not claim that synthetic data is free of all semantic shortcuts; rather, \pal{} makes leakage risks measurable and auditable at the field and identifier level, while Section~\ref{sec:experiments} tests whether current systems can solve the task from public observations alone.

This contract distinguishes \pal{} as a data-engineering benchmark rather than a prompt-only agent benchmark. The input is a heterogeneous collection with weak schema, noisy fields, and hidden joins; the output is a structured integration result with provenance. Section~\ref{subsec:characterization} quantifies the resulting difficulty profile.

\section{Evidence Compiler and Benchmark Characterization}
\label{sec:compiler}

\begin{figure*}[t]
\centering
\includegraphics[width=0.85\textwidth]{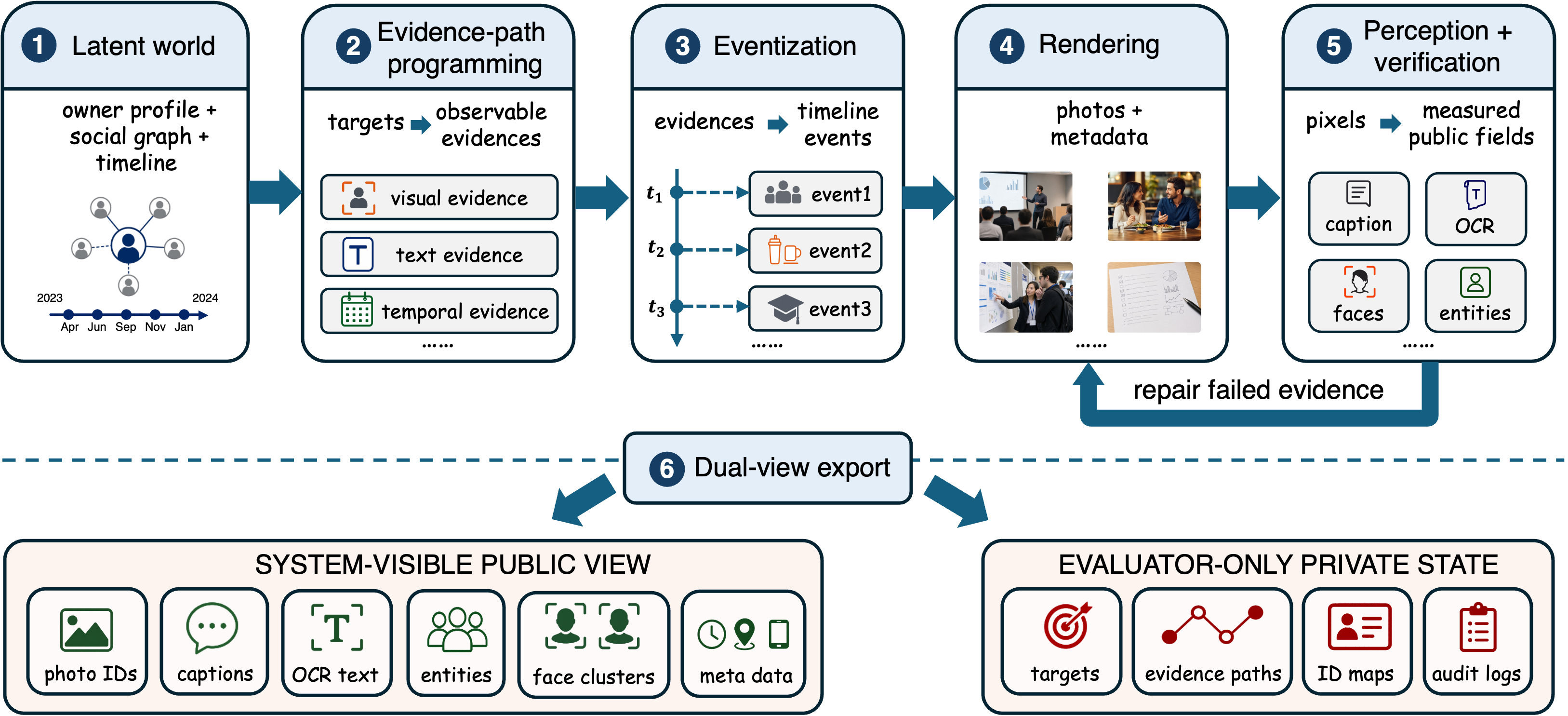}
\caption{The \pal{} evidence compiler. A latent private world is compiled into evidence paths and eventized photo plans; a fixed renderer produces pixels, perception and verification rebuild observable public fields, and dual-view export separates system-visible public records from evaluator-only targets, evidence paths, ID maps, and audit logs.}
\label{fig:pipeline}
\end{figure*}

The construction pipeline in Fig.~\ref{fig:pipeline} turns latent private worlds into audited public albums. We describe the compiler first, then quantify the resulting benchmark.

\subsection{Compiler Stages}
\label{subsec:compiler_stages}

\noindent\textbf{Latent world synthesis.}
For each user, the compiler creates a private owner profile (occupation, city, background, hobbies, routine, education, family structure, personality), a social graph (recurring people with names, relations, categories, demographics, personas), and a 12--24 month timeline. Diversity-aware sampling produces 39 occupations, 22 cities, 13 cultural backgrounds, 7 archetypes, and 6 social categories. We summarize each dimension by the \emph{normalized entropy} $\widetilde{H}(X) = H(X)/\log_2 K$, where $H(X)$ is the Shannon entropy of the empirical distribution over the $K$ values observed in the 50 users; $\widetilde{H}\in[0,1]$, with $\widetilde{H}=1$ at uniform coverage. The release attains $\widetilde{H} = 0.985$ (occupation), $0.956$ (archetype), $0.943$ (city), and $0.943$ (cultural background).

\smallskip\noindent\textbf{Evidence-path programming.}
Each target receives a private evidence path specifying modalities, observations, key photos, and reasoning. Difficulty reflects alignment complexity: easy paths use direct co-occurrence (face and name in one photo); medium paths require same-event alignment; hard paths require cross-event or cross-month alignment, e.g., linking a name in a screenshot to a recurring face months later.

\smallskip\noindent\textbf{Eventization, rendering, and verification.}
Evidence photos are embedded in coherent events---trips, meetings, birthdays, commutes, meals, classes, family gatherings, ambient scenes---preventing proof photos from appearing as isolated inserts and creating coherent distractor context (2,947 planned events; 4,869 key-evidence photos). The compiler renders all anchor portraits, key-evidence, event-scene, and ambient photos with Nano Banana 2. This renderer is a core construction dependency: subject consistency, instruction following, readable text rendering, and scene plausibility determine whether programmed face, OCR, relation, and event-context evidence survives rendering and can be measured. The rendered images are then measured through face detection and matching~\cite{taigman2014deepface,schroff2015facenet}, captioning~\cite{li2023blip2,liu2024llava,chen2023minigpt4}, OCR, and entity extraction. Public records store \emph{measured} content rather than private generation intent: if a generated image was intended to show a name badge but OCR cannot read it, the public record does not silently contain the intended name. For requirement-bearing photos, a verifier compares measured outputs against private requirements and triggers refinement when required evidence is missing. Across 38,378 manifest photos, 22,516 pass directly, 1,183 are fixed through refinement, 14,636 ambient photos have no requirement, and 43 remain in unverified/error states (Table~\ref{tab:quality}), with no confirmed failures. The dual-view export then drops manifest photos that fail the public/private invariants of \S\ref{subsec:separation} (the 43 unresolved photos and 1,676 records that fail at least one separation check), yielding the 36,659 agent-visible public records of \S\ref{subsec:release}.

\smallskip\noindent\textbf{Dual-view export.}
The public exporter rewrites photo IDs, remaps face IDs, strips private fields, and emits the public album view; the private exporter emits the evaluator view with targets, identifier maps, and audit metadata. Automated checks reject leakage of private IDs, photo roles, event IDs, image paths, verification status, owner labels, dangling face references, or invalid target references. Evidence compilation thus inverts post-hoc annotation: the inference route is created first, and observations are rendered to realize it; perception and verification prevent the compiler from trusting its intent. \pal{} combines constructive ground truth with observational public data---the evaluator knows why a target should be inferable, while the agent sees only what perception measured.

\subsection{Quality Control and Audit Outcomes}
\label{subsec:quality}

Table~\ref{tab:quality} summarizes construction-time checks. The compiler distinguishes intended from measured evidence and refines requirement-bearing photos that fail perception; ambient photos have no target-level requirement and are not forced through verification. The final export audit then checks the public/private separation contract per user.

\begin{table}[t]
\caption{Construction checks and export-audit outcomes. Construction-stage rows sum to manifest photos; the export audit yields the released public records.}
\label{tab:quality}
\centering
\scriptsize
\begin{tabular}{@{}lrr@{}}
\toprule
\textbf{Check} & \textbf{Count} & \textbf{Interpretation} \\
\midrule
Passed requirement check & 22,516 & Evidence realized directly \\
Fixed by refinement & 1,183 & Re-rendered and verified \\
Skipped ambient photos & 14,636 & No explicit requirement \\
Unverified / error / failed & 43 & 0 unrepaired failures \\
\midrule
Manifest photos (construction) & 38,378 & Sum of the four rows above \\
Dropped at export audit & 1,719 & 43 unresolved + 1,676 separation-fail \\
Released public records & 36,659 & Final agent-visible album \\
\midrule
Users passing export audit & 50/50 & Public/private contract holds \\
\bottomrule
\end{tabular}
\end{table}

These checks are evidence-realization audits, not only image-quality filters: a target is exported only when rendered pixels support the required public measurements, and missing requirement-bearing evidence is either repaired or excluded. They reduce the risk of generator-metadata leakage in two ways: representationally, public fields are rebuilt from perception outputs rather than copied from private prompts; structurally, private identifiers and target objects are removed and remapped before release. We discuss the resulting validity boundary in Section~\ref{subsec:why_synthetic}.

\subsection{Released Corpus}
\label{subsec:release}

The released benchmark contains 50 users, 36,659 agent-visible public records, 2,799 evaluation targets, 840 owner fact atoms, 653 social persons, and 703 public face clusters. Public album size averages 733 photos/user (range 563--1{,}000); target count averages 56/user (range 34--76). Public observation coverage is high: captions, timestamps, coarse location, and device metadata appear for all public records; OCR text in 82.4\%; text entities in 64.2\%; visible faces in 50.3\%.

\subsection{A Representative Evidence Chain}
\label{subsec:case}

Consider a hard social target: a recurring public face cluster $f^\star$ must be named and linked to the owner as a partner. The evidence path might contain four public observations. In month 3, a reservation screenshot contains the name ``Darius Campbell'' but no face. In month 5, the same owner appears at dinner with $f^\star$, but the caption contains no name. In months 7--12, $f^\star$ recurs in family and weekend contexts, creating a temporal relation pattern. In month 10, a birthday card contains visible text ``Happy Birthday Darius'' with $f^\star$ present. No single photo is sufficient; the target is supported by a chain that joins name strings to faces through event proximity, faces to relation categories through recurrence, and owner facts to public citations through repeated evidence---structure closer to an evidence graph than to a flat caption list.

\begin{figure}[t]
\centering
\includegraphics[width=\linewidth]{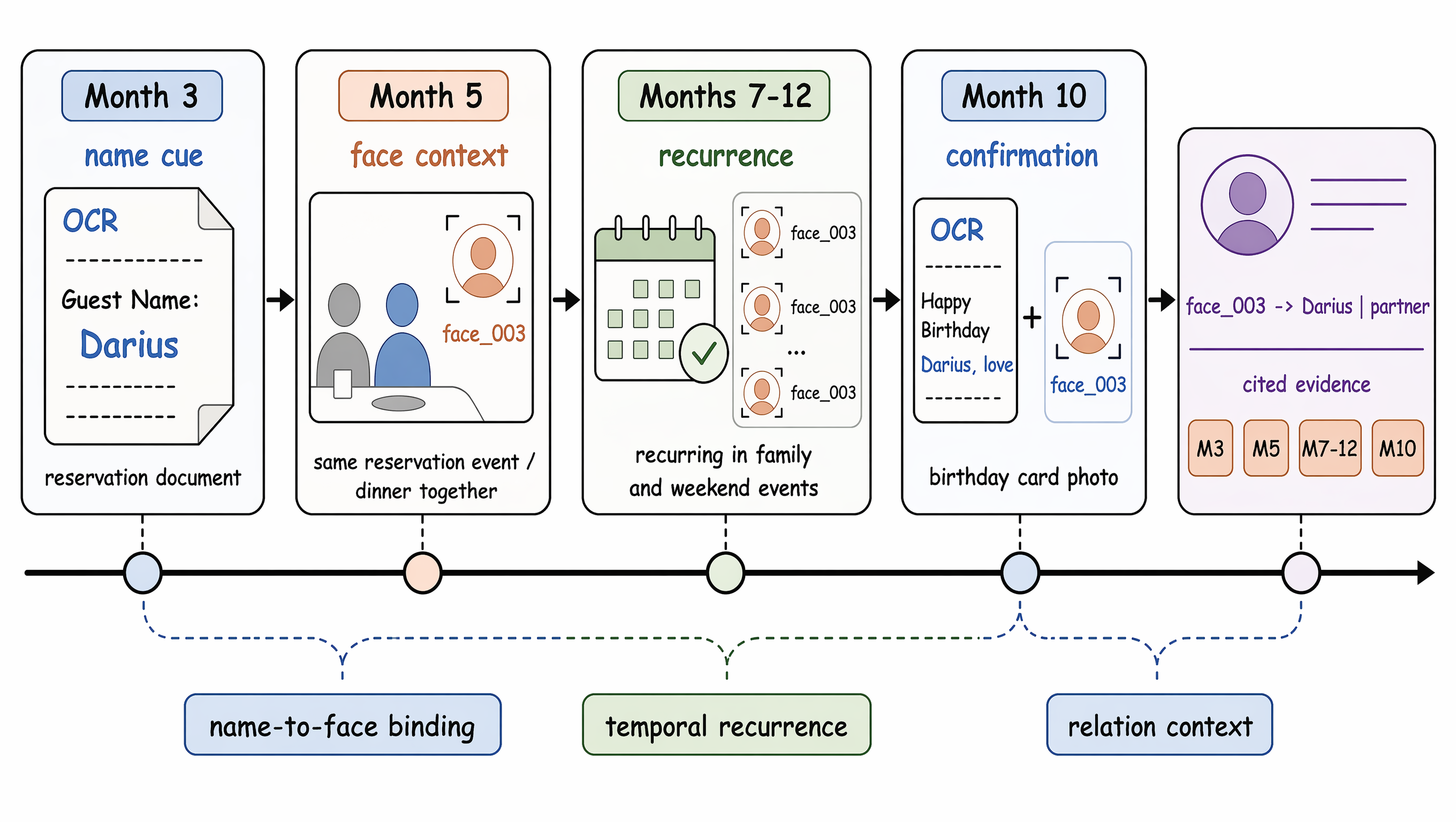}
\vspace{-20px}
\caption{Representative evidence chain for a hard social target (illustrative; all names and content are synthetic). Naming and relating \texttt{face\_003} requires joining an OCR name cue, same-event face context, recurring family/weekend appearances, and a text--face confirmation; the final row cites the supporting public records.}
\label{fig:evidence_chain}
\end{figure}

\subsection{Characterization Claims}
\label{subsec:characterization}

\begin{figure*}[t]
\centering
\includegraphics[width=0.9\textwidth]{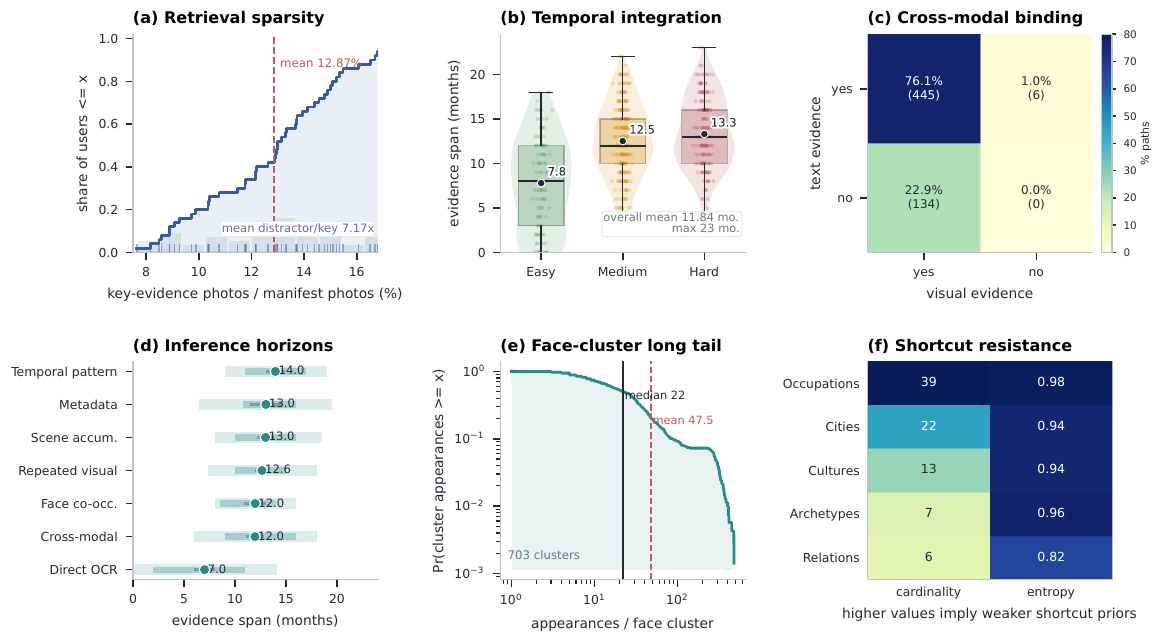}
\vspace{-15px}
\caption{Benchmark characterization dashboard. Panels show (a) sparse key evidence, (b) month-scale evidence spans, (c) visual-text cross-modal paths, (d) inference horizons by reasoning type, (e) long-tailed face-cluster appearances, and (f) diversity/entropy statistics that reduce shortcut priors.}
\label{fig:characterization}
\vspace{-10px}
\end{figure*}

Fig.~\ref{fig:characterization} and Table~\ref{tab:stress_profile} summarize five properties that explain why \pal{} stresses data-engineering capabilities rather than isolated visual recognition.

\smallskip\noindent\emph{(1) Needle-in-haystack retrieval.} Key evidence is only 12.87\% of public records on a per-user mean (12.69\% pooled), giving a per-user mean distractor/key ratio of 7.17$\times$ (Table~\ref{tab:stress_profile}); the per-user mean exceeds the simple corpus inversion $\approx 6.77\times$ because user albums vary in target density. The query is latent---the agent must infer what to search for---making this a discovery problem, not a filtering one.

\smallskip\noindent\emph{(2) Structural temporal integration.} Owner-fact evidence spans 11.84 months on average and 3.89 distinct evidence months per fact; restricting attention to any single calendar month would miss almost all key photos.

\smallskip\noindent\emph{(3) Cross-modal binding is the norm.} Owner-fact targets are organized into 585 cross-photo evidence paths; 76.1\% require both face and text. Text alone retrieves some owner facts but cannot bind identities; faces alone cluster recurring people but cannot name or explain them.

\smallskip\noindent\emph{(4) Long-tailed social identity.} The 653 social persons span friends (225), family (224), colleagues (101), classmates (49), neighbors (37), and others (17). Face appearances are heavy-tailed (mean 47.49, median 22, range 1--477); hard identities require cross-event recurrence.

\smallskip\noindent\emph{(5) Diversity against shortcut priors.} Using the normalized entropy of \S\ref{subsec:compiler_stages} we obtain $\widetilde{H} = 0.985$ (occupation), $0.956$ (archetype), and $0.943$ for both city and cultural background, discouraging corpus-level shortcuts.

\begin{table}[t]
\caption{Benchmark stress profile.}
\label{tab:stress_profile}
\centering
\scriptsize
\begin{tabular}{@{}lrl@{}}
\toprule
\textbf{Property} & \textbf{Value} & \textbf{Primary pressure} \\
\midrule
Key evidence fraction & 12.87\% & Sparse signal \\
Distractor/key ratio & 7.17$\times$ & Precision under clutter \\
Mean evidence span & 11.84 months & Temporal integration \\
Distinct months/fact & 3.89 & Cross-month aggregation \\
Cross-modal owner paths & 76.1\% & Face-text binding \\
Max face appearances & 477 & Long-tail recurrence \\
\bottomrule
\end{tabular}
\end{table}

\subsection{Construct-Coverage Audit on Real Private Albums}
\label{subsec:real_album_audit}

A construct-validity question for any synthetic benchmark is whether it evaluates structures that exist outside the generator. We address this through a privacy-preserving audit on real private albums that establishes \emph{construct coverage} rather than distributional equivalence: real private albums do contain the evidence structures that \pal{} controls and evaluates, but those structures are usually mundane and privacy-sensitive, making public release of real benchmark albums infeasible.



\noindent\textbf{Protocol.}
The audit comprises two locally-run instruments. \emph{(a) Coverage scan:} ten volunteers each inspected 200 stratified-random photos sampled by month and event-type bucket from their own private albums, reporting only aggregate, value-redacted statistics. \emph{(b) Fact-chain enumeration:} each participant additionally enumerated 12 value-redacted profile facts about themselves and tagged each fact, against a shared codebook, with the modalities, photo count, and privacy sensitivity of its supporting evidence chain. The procedures, codebook, and consent text were reviewed under the institutional human-subjects protocol covering anonymized self-audit; no photos, names, exact locations, screenshots, or raw metadata left participant devices. Table~\ref{tab:real_album_audit} reports participant-level coverage rates and fact-chain counts; the 10-participant sample tests \emph{construct presence} but does not support distributional inference, so we limit downstream claims to construct-coverage statements (Section~\ref{subsec:why_synthetic}).

\begin{table}[t]
\caption{Real-album evidence audit ($N=10$). $a/b$: participant coverage; $x/120$: redacted fact-chain counts over $10\times 12$ facts; right: linked \pal{} / \paltrace{} elements.}
\label{tab:real_album_audit}
\centering
\scriptsize
\begin{tabular}{@{}p{0.28\linewidth}p{0.38\linewidth}p{0.24\linewidth}@{}}
\toprule
\textbf{Construct} & \textbf{Audit result} & \textbf{Design link} \\
\midrule
Longitudinal content & 10/10 span $\geq$2 years; median 72\% non-public-shareable & Longitudinal setting \\
Recurring identities & 10/10 contain recurring non-owner people; median 11 identities & PIR; identity anchoring \\
Cross-photo chains & 101/120 redacted facts require $\geq$2 photos; median 4 photos/fact & Evidence chains; EFS \\
Multimodal support & 79/120 facts require $\geq$2 modalities & Face+OCR+time/location \\
Text traces & 10/10 contain readable text; median 17\% text-bearing photos & OCR/document evidence \\
Privacy barrier & 94/120 fact chains rely on sensitive evidence; median 21\% public-shareable & Need for synthetic benchmark \\
\bottomrule
\end{tabular}
\vspace{-10px}
\end{table}

\noindent\textbf{Findings.}
All six constructs that \pal{} stresses occur in real private albums: recurring non-owner identities, multi-photo fact chains, multimodal support, text traces, longitudinal span, and privacy-sensitive evidence. The audit also bounds what should \emph{not} be claimed. It is a construct-coverage study, not a distributional sample: real-album median 11 recurring identities/user is close to \pal{}'s 13.06, but text evidence is much sparser in the audit (17\% text-bearing photos) than in \pal{} (82.4\%). We therefore characterize \pal{} as a controlled stress test with an upper-bound text channel, not as a simulator of real-album frequencies. Where real albums have sparser text, identity binding shifts further onto face/temporal/event-context evidence---the modalities whose removal collapses identity resolution in our face-off ablation (Section~\ref{subsec:rq3}). The audit establishes that the modeled structures are not generator artifacts; it does not establish distributional equivalence (see Section~\ref{subsec:why_synthetic}).

\section{Evaluation Protocol and Reference Systems}
\label{sec:evaluation}

\subsection{Metric Suite}
\label{subsec:metrics}


The evaluation protocol uses seven metrics---OFR, PIR, PIR-hard, PRR-ID, EFS, ECE, and \nllm{}---whose definitions are given by the prose and equations in this subsection. Metrics are macro-averaged over users unless otherwise stated. Semantic matching and evidence judgments use a fixed judge with cached outputs and identical scoring prompts across systems. The judge role is fixed to Qwen3.6-35B-A3B at temperature 0; the value-match prompt, the EFS 0/0.5/1 rubric, and the cached judge outputs accompany the supplementary artifact report so that all OFR/EFS values can be reproduced from the cache or recomputed from scratch.


For owner facts, a prediction matches a target if it conveys the same factual content without adding unsupported specifics. Relation and category for social persons are scored only after correct identity binding, avoiding relation credit on the wrong identity. EFS is judged on cited public photo IDs and claim text; ECE is pooled with binned confidences; \nllm{} counts reconstruction calls only.

\smallskip\noindent\textbf{Judge and metric scope.}
The semantic judge is used as a deterministic, cached evaluator for target matching and citation support, not as an oracle for all possible profile statements. A second-judge analysis in Section~\ref{subsec:rq6_robust} tests whether method ordering is robust to judge choice. OFR is a target-recovery metric rather than a global precision estimate over arbitrary extra claims outside the hidden target inventory; open-world equivalence over unconstrained profile statements is outside the benchmark contract. We therefore interpret OFR jointly with identity-conditioned metrics, EFS, ECE, emitted-row counts, WrongFaceRate, and the error analysis rather than as a standalone measure of profile quality.


Let $\mathcal{U}$ be users and $\mathcal{T}^{of}_u$ owner-fact targets:
\begin{equation}
\text{OFR} =
\frac{1}{|\mathcal{U}|}\sum_{u \in \mathcal{U}}
\frac{1}{|\mathcal{T}^{of}_u|}
\sum_{t \in \mathcal{T}^{of}_u}\text{match}(t,\hat{O}_u).
\end{equation}
For identity, let $S_u$ be the set of ground-truth social-person targets for user $u$, $D_u$ the number of those identities a system discovers anywhere in its predicted profile, and $B_u$ the number it additionally binds to the correct public face cluster. Then $\text{IDR}_u = D_u/|S_u|$ and $\text{IBR}_u = B_u/D_u$, and PIR is their harmonic mean (set to 0 when $D_u=0$), macro-averaged over users. The harmonic mean is an F1-style combination that prevents credit for listing many names without binding them or for binding only a few easy people. PIR-hard restricts $S_u$ to hard persons; PRR-ID conditions on $B_u$ and measures relation accuracy among correctly bound persons. EFS averages a 0/0.5/1 support score from the fixed semantic judge over matched claims. ECE pools all evaluated targets into 10 equal-width confidence bins $\{B_b\}_{b=1}^{10}$ and uses the standard form $\text{ECE} = \sum_{b} (n_b/N)\,|\text{acc}(b)-\text{conf}(b)|$, where $n_b=|B_b|$, $N$ is the total pooled target count, and $\text{acc}(b)/\text{conf}(b)$ are the bin-mean correctness and bin-mean confidence.

\subsection{Reference Systems}
\label{subsec:systems}
We evaluate seven primary systems plus two compute-matched diagnostics. Across them, the primary design axis is where identity state resides: in rules, nowhere explicit, retrieved context, in-context summaries, tool traces, timeline-centric aggregation, or a state-separated composition. The \emph{no-LLM evidence heuristic} aggregates face clusters, OCR entities, and temporal co-occurrence by rule, testing how far structured signals go without generative reasoning. The \emph{text-only LLM extractor} reads captions, OCR, entities, timestamps, and metadata but no face clusters---a control for text evidence rather than a social-identity system. \emph{Multimodal RAG} retrieves on text/face/metadata features and uses one bounded LLM composition call~\cite{lewis2020rag,gao2023retrievalaugmentedmultimodal}. The \emph{long-context multimodal LLM} serializes compressed album evidence into a long context and produces the full profile in one pass, testing whether context length substitutes for structured retrieval and identity state. The \emph{compute-scaled long-context diagnostic} adds six flat summary passes plus a final composition over information-dense records (bounded per-photo serialization, not a whole-album dump), matching \paltrace{}'s call budget without its frozen identity table. The \emph{generic tool-use agent}~\cite{yao2023react,shinn2023reflexion} uses one planning call to issue text-search and face-inspection actions and one composition call; we also run a 7-call diagnostic variant with fixed exploration calls (planning, owner/identity/relation/temporal notes, evidence selection, composition), still without a frozen identity table. The \emph{adapted prior/lifelog baseline} groups photos into month-location episodes, summarizes recurring events and faces, and emits the same schema, adapting personal-media QA assumptions without explicit queries. \emph{\paltrace{}} (Personal Album Traceable Reconstruction via Anchored Cross-modal Evidence) is our reference framework: it separates state that should remain stable (identity) from facts that benefit from type-open expansion. All systems consume the same multimodal public album view (text-only hides faces) and emit the same open structured output, discovering recoverable claims rather than answering provided questions.

\begin{figure*}[t]
\centering
\includegraphics[width=0.9\textwidth]{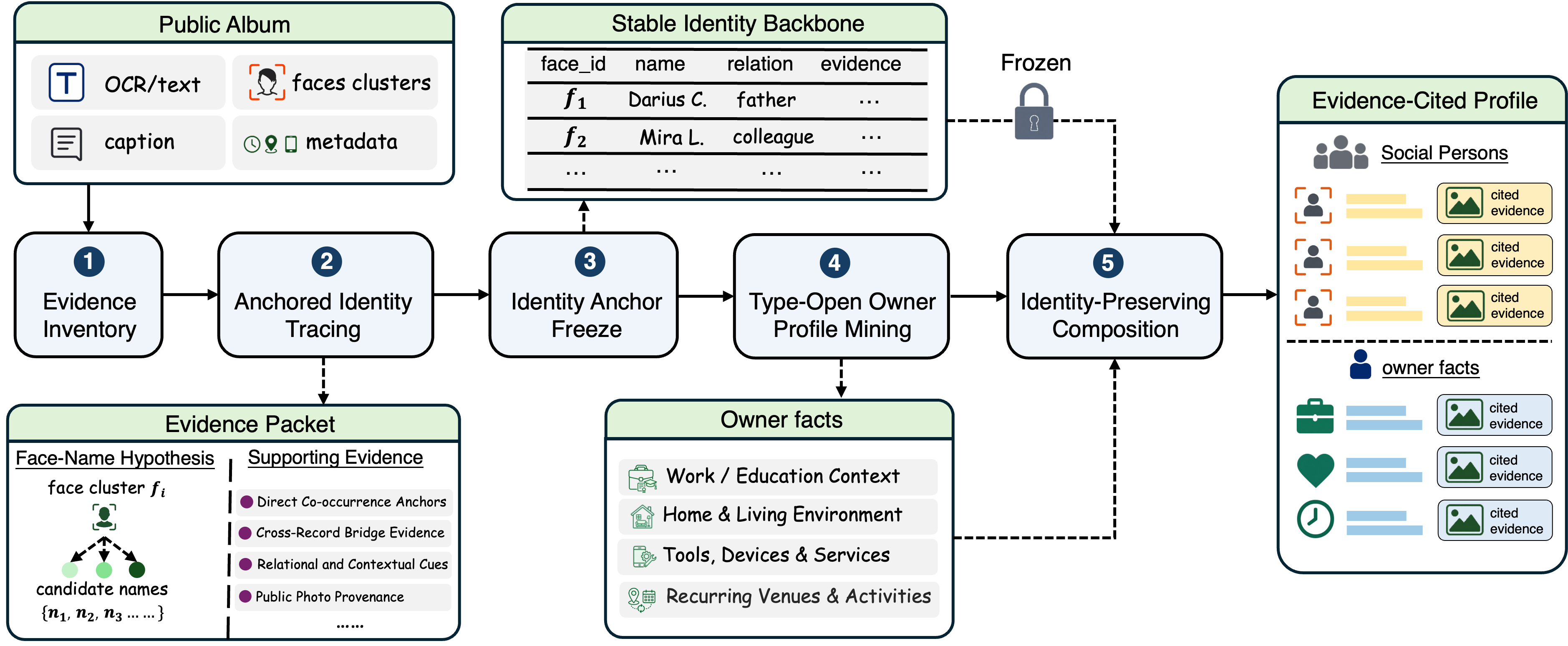}
\caption{\paltrace{} state-separated reconstruction. Public album fields are converted into evidence packets; anchored identity tracing freezes a stable identity backbone, while type-open owner-profile mining extracts owner facts from the same evidence inventory. Identity-preserving composition copies social-person rows from the frozen backbone and owner facts from the owner-mining output, blocking identity write-back and enforcing Eq.~\ref{eq:composer}.}
\vspace{-10px}
\label{fig:paltrace}
\end{figure*}

\begin{algorithm}[t]
\caption{\paltrace{}: state-separated reconstruction with identity-preserving composition.}
\label{alg:paltrace}
\scriptsize
\begin{algorithmic}[1]
\REQUIRE Public album $A_u$
\ENSURE Evidence-cited profile $\hat{Y}_u$ satisfying Eq.~\ref{eq:composer}
\STATE \begin{minipage}[t]{0.97\linewidth}\raggedright
\textbf{Stage 1 (deterministic).}
$G \leftarrow \textsc{BuildEvidenceInventory}(A_u)$\\[-1pt]
{\color{gray}\textit{// face clusters, OCR, entities, time/location cues}}
\end{minipage}
\STATE \begin{minipage}[t]{0.97\linewidth}\raggedright
\textbf{Stage 2 (stable identity run).}
$\hat{P}^{\textrm{stab}}_u \leftarrow \textsc{AnchoredIdentityTracing}(G)$\\[-1pt]
{\color{gray}\textit{// candidate generation, joint assignment, repair, delayed binding}}
\end{minipage}
\STATE \begin{minipage}[t]{0.97\linewidth}\raggedright
\textbf{Stage 3 (guarded owner mining over $G$).}\\[-1pt]
\quad $\hat{O}^{\textrm{own}}_u \leftarrow \textsc{TypeOpenOwnerMining}(G)$\\[-1pt]
\quad $\textsc{PIIGuard}$: drop emails, phone numbers, street addresses\\[-1pt]
\quad $\textsc{StructGuard}$: drop owner-name restatements and named-friend lists\\[-1pt]
\quad $\textsc{EvidGuard}$: drop disjunctive, hedged, or uncited claims
\end{minipage}
\STATE \textbf{Stage 4 (deterministic composer).}
\STATE \quad $\hat{Y}_u.\text{persons} \leftarrow \hat{P}^{\textrm{stab}}_u$
{\color{gray}\textit{// verbatim copy}}
\STATE \quad $\hat{Y}_u.\text{owner.facts} \leftarrow \hat{O}^{\textrm{own}}_u.\text{facts}$
{\color{gray}\textit{// owner facts only}}
\RETURN $\hat{Y}_u$ with public record citations
\end{algorithmic}
\end{algorithm}

\subsection{\paltrace{}: State-Separated Reconstruction}
\label{subsec:paltrace}

\smallskip\noindent\textbf{Thesis.}
Album reconstruction has two intertwined unknowns: \emph{who recurs} and \emph{what is true about the owner}. Long-context, RAG, and agent baselines infer both jointly. \paltrace{} instead instantiates a database-style principle: stabilize identity bindings as persistent entity state, then mine type-open owner facts under that state. Identity, once resolved from cross-modal evidence, becomes a constraint that later stages may consult but not rewrite; owner facts remain schema-open and evidence-driven. Findings 1b, 3, and 5 test the resulting predictions.

\smallskip\noindent\textbf{State-separation invariant.}
Let $A_u$ be the public album, $\hat{P}^{\textrm{stab}}_u$ the social-person rows produced by an identity stage, and $\hat{O}^{\textrm{own}}_u$ the owner-fact rows produced by an owner-mining stage. \paltrace{} defines the final profile as
\begin{equation}
\hat{Y}_u \;=\; \mathcal{C}\!\left(\hat{P}^{\textrm{stab}}_u,\;\hat{O}^{\textrm{own}}_u\right),
\label{eq:composer}
\end{equation}
where the composer $\mathcal{C}$ is a pure function that copies $\hat{P}^{\textrm{stab}}_u$ verbatim into the social-person section of $\hat{Y}_u$ and reads from $\hat{O}^{\textrm{own}}_u$ \emph{only its owner-fact field}---any identity-shaped output of the owner-mining stage is structurally discarded. This yields:

\smallskip\noindent\emph{Identity-Preservation Property.} \emph{For every album $A_u$ and every owner-mining realization $\hat{O}^{\textrm{own}}_u$, the social-person rows of $\hat{Y}_u$ depend only on $\hat{P}^{\textrm{stab}}_u$.}

\smallskip\noindent
The property is enforced by system structure, not prompt language: $\mathcal{C}$ ignores identity-shaped fields from $\hat{O}^{\textrm{own}}_u$ even if the owner-mining LLM produces them. Cost accounting follows the same separation, $\nllm{}_u = \nllm{}^{\textrm{stab}}_u + \nllm{}^{\textrm{own}}_u$. The closest data-management analogues are master-data management---one canonical entity table as the system of record~\cite{halevy2006data,christophides2021entityresolution}---and snapshot isolation---a downstream stage cannot mutate a frozen prior view.

\smallskip\noindent\textbf{Anchored identity tracing.}
The stable identity stage builds $\hat{P}^{\textrm{stab}}_u$ from the inventory $G$ (face clusters, OCR/entities, captions, time/location cues). It enumerates name candidates per public face cluster using same-photo co-occurrence, cross-photo event bridges, recurrence, and venue context, then applies bounded LLM adjudication for joint assignment, hard-block repair, and adaptive repair. Final binding is \emph{delayed} until text, face, and event evidence have been aggregated, mirroring entity-resolution practice~\cite{christophides2021entityresolution,mudgal2018deepmatcher,li2020ditto}. The goal is a stable backbone for PIR/PIR-hard/PRR-ID, not speculative person rows.

\smallskip\noindent\textbf{Type-open owner mining.}
Owner facts span hobbies, work settings, routines, devices, services, organizations, and household roles, so no fixed schema covers the space without forcing unsupported slots. Owner mining therefore emits in an \emph{open world} but rejects in a \emph{closed world}, using three guard classes:
\emph{(i) PII-leak guards} reject exact emails, phone numbers, and street addresses, which a benchmark with citations should not surface even when public photos contain them;
\emph{(ii) structural-placement guards} reject owner-name restatements and named-friend lists, since those rows belong to the identity backbone, not the owner-fact field;
\emph{(iii) evidence-quality guards} reject disjunctive claims (``A or B''), hedged claims (``may'', ``possibly''), and low-confidence claims without concrete public photo evidence.
The first two prevent leakage into the wrong section; the third makes the owner channel evaluable under EFS. The album, not a schema, decides what is recoverable.

\smallskip\noindent\textbf{Framework, not agent loop.}
Stage order is fixed and the planner sits outside the LLM. Each LLM call fills a typed slot, dispatched and composed by the framework, unlike ReAct, Reflexion, and AutoGen-style loops~\cite{yao2023react,shinn2023reflexion,wu2024autogen} where an LLM decides what to update. We therefore avoid joint identity/fact mining in one pass (the \emph{Fresh owner mining} ablation degrades PIR-hard from $0.268$ to $0.231$, Section~\ref{subsec:rq5_mechanism}) and avoid fixed owner schemas that would force unsupported slots.

\smallskip\noindent\textbf{Three falsifiable predictions.}
If state separation is useful rather than incidental:
\emph{(P1)} adding LLM calls under joint identity/owner reasoning should \emph{not} match \paltrace{}'s gains, and may \emph{degrade} identity binding by re-deriving bindings under fresh prompt context---tested in Section~\ref{subsec:rq1} (compute-scaled long-context and 7-call generic tool-use diagnostics);
\emph{(P2)} face evidence should be structurally necessary, since the stable identity stage cannot anchor without it---tested by the face-off ablation in Section~\ref{subsec:rq3};
\emph{(P3)} the gain should transfer across LLM backbones, since the framework rather than a single backbone implements the identity-state discipline---tested in Section~\ref{subsec:rq6_robust}.
The mechanism table gives the direct signature: \paltrace{}'s identity rows match the stable identity stage \emph{by construction}, while owner recovery and EFS gain $+9.1$ and $+2.9$ points and identity metrics resist the \emph{Fresh owner mining} perturbation.

\smallskip\noindent\textbf{Scope.}
\paltrace{} assumes \pal{}'s public interface: perception-derived face clusters with stable public identifiers and a public-only evidence surface. Settings without such identifiers need an upstream perceptual entity-resolution layer; the state-separation invariant is independent of that layer.

\section{Experiments}
\label{sec:experiments}

\subsection{Experimental Setup}
\label{subsec:setup}

All systems are evaluated on the 50-user \pal{} release. The official judge is a fixed semantic evaluator (Qwen3.6-35B-A3B) used for both fact matching and evidence support; the primary \paltrace{} run uses GPT-5.4 as the reconstruction backbone, which we pre-specified as the default before any backbone-robustness comparisons (Section~\ref{subsec:rq6_robust}) so that headline numbers are not selected to favor \paltrace{}. We use macro-averaging over users and paired bootstrap confidence intervals (CIs) computed with $10^4$ user-level resamples; EFS and ECE are evaluated only on matched claims with cited evidence. Each system receives the same public album view, produces the same output schema, and is not tuned on private targets. Validity probes (compute-scaled long-context, 7-call generic tool-use, caption-swap text-only, second-judge rescoring on a 10-user subset) make the comparison diagnostic rather than merely competitive.

\subsection{RQ1: Do Current Systems Solve PAL-Bench?}
\label{subsec:rq1}

Table~\ref{tab:main_results} reports the full comparison. No baseline dominates across metrics, and none saturates the benchmark. The strongest non-\paltrace{} OFR is the compute-scaled long-context diagnostic (0.4832), but its PIR is only 0.0753 because repeated summarization does not preserve stable face-name bindings; the 7-call generic tool-use agent likewise does not help (OFR/PIR/EFS = 0.2353/0.1509/0.1713). The strongest primary-baseline PIR is the deterministic heuristic (0.4047), but its EFS is only 0.2306 and OFR 0.3799. RAG and long-context systems exhibit a consistent split: PRR-ID is high once identity is correct, but PIR is low---\emph{binding people} is harder than labeling already-bound people.

\begin{table*}[t]
\caption{Main results on 50 users. Higher is better except ECE and \nllm{}; ``NA'' means undefined with no correctly bound identities. \paltrace{}'s identity rows equal the stable identity run by construction (Eq.~\ref{eq:composer}); see Table~\ref{tab:mechanism}.}
\label{tab:main_results}
\centering
\scriptsize
\begin{tabular}{@{}lrrrrrrr@{}}
\toprule
\textbf{System} & \textbf{OFR} & \textbf{PIR} & \textbf{PIR-hard} & \textbf{PRR-ID} & \textbf{EFS} & \textbf{ECE} & \textbf{\nllm{}} \\
\midrule
No-LLM evidence heuristic & 0.3799 & 0.4047 & 0.1756 & 0.5749 & 0.2306 & 0.3749 & 0.0000 \\
Text-only LLM extractor & 0.4041 & 0.0000 & 0.0000 & NA & 0.1088 & 0.0161 & 1.0000 \\
Multimodal RAG & 0.3580 & 0.2942 & 0.1573 & 0.9175 & 0.2213 & 0.1116 & 1.0000 \\
Long-context multimodal LLM & 0.4009 & 0.2336 & 0.1617 & 0.8446 & 0.1968 & 0.1075 & 1.0000 \\
Compute-scaled long-context & 0.4832 & 0.0753 & 0.0398 & 0.9861 & 0.2014 & 0.1387 & 7.0400 \\
Generic tool-use agent & 0.2560 & 0.3321 & 0.1405 & 0.8352 & 0.2039 & 0.1628 & 2.1000 \\
Generic tool-use (7-call diag.) & 0.2353 & 0.1509 & 0.0660 & 0.9062 & 0.1713 & 0.1858 & 7.0400 \\
Adapted prior/lifelog baseline & 0.3921 & 0.3900 & 0.2098 & 0.8328 & 0.2741 & 0.1528 & 1.0000 \\
\textbf{\paltrace{}} & \textbf{0.6057} & \textbf{0.4792} & \textbf{0.2684} & 0.8384 & \textbf{0.3764} & 0.3236 & 7.4400 \\
\bottomrule
\end{tabular}
\vspace{-10px}
\end{table*}

\noindent\textbf{Finding 1.} Album-scale profile reconstruction is not solved by context length or generic retrieval. \paltrace{} improves OFR by 20.5 points over one-call long-context and 24.8 over RAG, with EFS gains of 18.0 and 15.5. Paired bootstrap deltas against all six baselines (Table~\ref{tab:bootstrap}) keep OFR, PIR, and EFS intervals strictly above zero; non-crossing $10^4$-resample intervals bound one-sided $p$ by $\approx10^{-4}$~\cite{davison1997bootstrap}. The three pre-specified primary comparisons remain Holm-significant at $k{=}3$ ($p<3.0\times10^{-4}$). PIR-hard gains are positive but not always conclusive against one-call long-context and the adapted lifelog baseline.

\noindent\textbf{Finding 1b.} Extra LLM calls without identity-state discipline can degrade binding. Compute-scaled long-context adds six summarization passes: OFR rises from 0.4009 to 0.4832, but PIR drops from 0.2336 to 0.0753 and PIR-hard from 0.1617 to 0.0398. The 7-call tool-use diagnostic similarly drops from PIR/EFS 0.3321/0.2039 to 0.1509/0.1713. Since \paltrace{} spends 7.44 calls/user and gains, the evidence supports Prediction P1: \pal{} rewards \emph{state-disciplined} compute scaling, not raw call count.

\begin{table*}[t]
\caption{Paired bootstrap deltas for \paltrace{} minus selected baselines (95\% CIs over users).}
\label{tab:bootstrap}
\centering
\scriptsize
\begin{tabular}{@{}lrrrr@{}}
\toprule
\textbf{Baseline} & \textbf{$\Delta$OFR} & \textbf{$\Delta$PIR} & \textbf{$\Delta$PIR-hard} & \textbf{$\Delta$EFS} \\
\midrule
Long-context MM LLM & +0.2048 [0.1637, 0.2452] & +0.2456 [0.1811, 0.3117] & +0.1067 [-0.0085, 0.2168] & +0.1795 [0.1517, 0.2099] \\
Compute-scaled LC & +0.1225 [0.0854, 0.1603] & +0.4039 [0.3491, 0.4587] & +0.2286 [0.1299, 0.3225] & +0.1750 [0.1464, 0.2041] \\
Multimodal RAG & +0.2478 [0.2087, 0.2842] & +0.1850 [0.1109, 0.2586] & +0.1111 [0.0047, 0.2132] & +0.1550 [0.1233, 0.1886] \\
Adapted lifelog baseline & +0.2137 [0.1793, 0.2485] & +0.0892 [0.0186, 0.1598] & +0.0586 [-0.0413, 0.1571] & +0.1022 [0.0696, 0.1372] \\
Generic tool-use agent & +0.3497 [0.3125, 0.3857] & +0.1471 [0.0888, 0.2074] & +0.1279 [0.0407, 0.2161] & +0.1724 [0.1421, 0.2038] \\
Generic tool-use 7-call & +0.3705 [0.3250, 0.4158] & +0.3283 [0.2700, 0.3871] & +0.2024 [0.1171, 0.2879] & +0.2051 [0.1772, 0.2331] \\
\bottomrule
\end{tabular}
\vspace{-10px}
\end{table*}

The table also clarifies what \paltrace{} does \emph{not} solve: ECE is not best, and PIR-hard remains 0.2684. It reaches IDR 0.481 and IBR 0.559 with WrongFaceRate 0.208, emitting 13.06 person rows/user and 12.24 named rows/user, so its PIR gain reflects a larger discovered-and-bound identity set rather than an unbounded list. Two patterns stand out: PRR-ID is high for RAG (0.9175), long-context (0.8446), and compute-scaled long-context (0.9861) despite weak PIR, meaning relation labeling is easier once binding is correct; text-only extraction has competitive OFR and ECE but zero PIR, so owner summarization alone does not solve \pal{}.

\subsection{RQ2: How Does Difficulty Affect Performance?}
\label{subsec:rq2}
Difficulty strata expose which capabilities break first. Table~\ref{tab:difficulty} shows that \paltrace{}'s identity resolution degrades sharply with difficulty (PIR 0.37/0.26/0.17 across easy/medium/hard), while OFR follows a non-monotonic pattern: hard owner facts are not always below medium ones because difficulty reflects evidence-alignment complexity rather than rarity. EFS tracks PIR more closely than OFR.

\begin{table}[t]
\caption{Difficulty-stratified \paltrace{} results, with hard-target PIR for RAG and long-context baselines.}
\label{tab:difficulty}
\centering
\scriptsize
\begin{tabular}{@{}lrrr@{}}
\toprule
\textbf{Metric/System} & \textbf{Easy} & \textbf{Medium} & \textbf{Hard} \\
\midrule
\paltrace{} OFR & 0.7474 & 0.4948 & 0.5625 \\
\paltrace{} PIR & 0.3704 & 0.2610 & 0.1657 \\
\paltrace{} EFS & 0.4967 & 0.3345 & 0.2938 \\
\midrule
RAG PIR & -- & -- & 0.0888 \\
Long-context PIR & -- & -- & 0.1006 \\
\bottomrule
\end{tabular}
\vspace{-10px}
\end{table}

\noindent\textbf{Finding 2.} Hard targets are hard because they require identity-aligned evidence over time. Multimodal RAG reaches 0.0888 hard PIR and long-context 0.1006; \paltrace{} improves hard identity resolution but still binds fewer than one in five hard persons. \pal{} is non-saturated, and hard social identity binding is the benchmark's hardest axis.

\subsection{RQ3: Which Evidence Sources Matter?}
\label{subsec:rq3}

We run field-level ablations by masking text, face, or metadata evidence before reconstruction. The ablation is intentionally field-based: it measures the value of public evidence channels, not pixel-level robustness.

\begin{table}[t]
\caption{Field-level modality ablations for \paltrace{} and long-context baselines.}
\label{tab:ablation}
\centering
\scriptsize
\setlength{\tabcolsep}{2.2pt}
\begin{tabular}{@{}llrrrrr@{}}
\toprule
\textbf{Sys.} & \textbf{Input} & \textbf{OFR} & \textbf{PIR} & \textbf{PIR-h} & \textbf{EFS} & \textbf{$\Delta$EFS} \\
\midrule
\multirow{4}{*}{\paltrace{}}
& Full & 0.6057 & 0.4792 & 0.2684 & 0.3764 & 0.0000 \\
& Text-off & 0.3411 & 0.0000 & 0.0000 & 0.1774 & -0.1990 \\
& Face-off & 0.6155 & 0.0000 & 0.0000 & 0.1605 & -0.2159 \\
& Meta-off & 0.5919 & 0.3400 & 0.1694 & 0.3027 & -0.0737 \\
\midrule
\multirow{4}{*}{Long-ctx}
& Full & 0.4009 & 0.2336 & 0.1617 & 0.1968 & 0.0000 \\
& Text-off & 0.1420 & 0.0029 & 0.0000 & 0.1212 & -0.0756 \\
& Face-off & 0.3998 & 0.0000 & 0.0000 & 0.1094 & -0.0874 \\
& Meta-off & 0.3829 & 0.2339 & 0.1761 & 0.1911 & -0.0057 \\
\bottomrule
\end{tabular}
\vspace{-10px}
\end{table}

\noindent\textbf{Finding 3.} Text carries much of the owner-fact signal, but faces are structurally required for identity. Removing text from \paltrace{} drops OFR by 26.5 points and eliminates PIR. Removing faces leaves OFR superficially high (0.6155) because owner facts can still be mined from text, but PIR and PIR-hard become zero and EFS drops by 21.6 points---fact recovery alone can therefore mask identity failure. Metadata has a smaller but nontrivial effect, especially on identity disambiguation (PIR drops from 0.4792 to 0.3400).

A caption-swap shortcut probe tests whether public captions act as a hidden oracle. We preserve target user IDs, timestamps, metadata, OCR, faces, and evaluation ground truth, but replace captions and caption-sourced text entities with cross-user donor content matched by month, text density, and face count. The text-only extractor degrades from 0.4041 to 0.3498 OFR and from 0.1088 to 0.0932 EFS, with PIR remaining 0 in both settings (paired bootstrap deltas $-0.0543$ OFR $[-0.0859,-0.0230]$ and $-0.0156$ EFS $[-0.0237,-0.0076]$). Captions therefore carry real public signal but are not a single hidden answer channel: OCR, metadata, and non-caption fields still support recoverable owner facts.

The face-off result is the most informative ablation. OFR slightly increases because the system can focus on text-rich owner clues without allocating effort to social identity, but the evidence-cited profile is worse: social rows disappear, identity-conditioned metrics collapse, and cited evidence supports a narrower subset of claims. Album reconstruction therefore cannot be evaluated as profile summarization alone---a system that summarizes the owner while dropping social identities does not solve the benchmark.

\subsection{RQ4: Robustness to Real-Album Messiness}
\label{subsec:rq_reality}

Real personal albums tend to have shorter or sparser captions, missed OCR, missing geotags, and missed face detections; \pal{}'s public fields, though perception-derived, are still cleaner along these axes (notably OCR: real-album median 17\% vs.\ \pal{} 82.4\%). The Realism Stress Suite tests whether our conclusions depend on this gap by perturbing only agent-visible public fields while preserving user IDs, photo IDs, the face-ID namespace, and the private evaluator view. The most informative single setting is the \emph{combined-mild} perturbation: caption truncation to median 51\% characters, 25\% OCR-photo dropout, 30\% location sparsification, and 10\%/2\% non-owner/owner face-appearance dropout. We re-run \paltrace{} and the long-context multimodal LLM baseline under the same official judge protocol and report paired deltas against the unperturbed inputs.

\begin{table*}[t]
\caption{Realism Stress Suite: combined-mild vs.\ full input.}
\label{tab:reality}
\centering
\scriptsize
\begin{tabular}{@{}llrrrrrrr@{}}
\toprule
\textbf{System} & \textbf{Input} & \textbf{OFR} & \textbf{PIR} & \textbf{PIR-hard} & \textbf{PRR-ID} & \textbf{EFS} & \textbf{ECE} & \textbf{\nllm{}} \\
\midrule
\multirow{3}{*}{\paltrace{}}
& Full & 0.6057 & 0.4792 & 0.2684 & 0.8384 & 0.3764 & 0.3236 & 7.4400 \\
& Combined-mild & 0.5845 & 0.4043 & 0.2142 & 0.8149 & 0.3183 & 0.3208 & 7.3600 \\
& $\Delta$ (mild $-$ full) & $-0.021^{*}$ & $-0.075^{**}$ & $-0.054$ & $-0.019$ & $-0.058^{**}$ & $-0.000$ & $-0.08$ \\
\midrule
\multirow{3}{*}{Long-context}
& Full & 0.4009 & 0.2336 & 0.1617 & 0.8446 & 0.1968 & 0.1075 & 1.0000 \\
& Combined-mild & 0.4237 & 0.4282 & 0.2334 & 0.8662 & 0.3135 & 0.1585 & 1.0000 \\
& $\Delta$ (mild $-$ full) & $+0.023$ & $+0.195^{**}$ & $+0.072$ & $+0.019$ & $+0.117^{**}$ & $+0.077^{**}$ & $0.00$ \\
\bottomrule
\multicolumn{9}{l}{\scriptsize $^{*}$ 95\% CI excludes zero; $^{**}$ paired-bootstrap $p < 0.005$. Long-context PIR/EFS gain under degradation is paired with $+0.077$ ECE.} \\
\end{tabular}
\vspace{-10px}
\end{table*}

\noindent\textbf{Finding 4.} \paltrace{} degrades smoothly under real-album-inspired public-field perturbation: $-0.021$ OFR, $-0.075$ PIR, $-0.058$ EFS (OFR/PIR/EFS intervals exclude zero; PIR-hard and PRR-ID overlap zero; ECE unchanged). The long-context baseline's apparent improvement under degradation ($+0.195$ PIR, $+0.117$ EFS) is a calibration artifact: ECE worsens by $+0.077$, BlankNameRate (the fraction of target faces with no predicted name) rises from $0.42$ to $0.63$, IDR drops from $0.48$ to $0.32$, and IBR rises from $0.42$ to $0.78$---the baseline discovers fewer identities but over-commits on what remains. Under matched combined-mild input, \paltrace{} retains $+0.161$ OFR over long-context (CI $[+0.123,+0.198]$, $p<0.001$); PIR and EFS converge. The capability ordering of Section~\ref{subsec:rq1} therefore holds qualitatively under public-field degradation, with absolute scores moving in the direction the audit anticipates.

\subsection{RQ5: What Explains PAL-TRACE's Gains?}
\label{subsec:rq5_mechanism}
We probe the state-separation invariant of Eq.~\ref{eq:composer} by holding the stable identity stage fixed and varying only what the owner-mining stage is allowed to write. Two variants instantiate this: \emph{Stable evidence-chain} reports $\hat{P}^{\textrm{stab}}_u$ alone with no type-open owner expansion (the lower envelope of the framework with no owner gain); \emph{Fresh owner mining} runs the type-open owner stage but lets it emit social-person rows alongside owner facts, removing the composer's discard step (the natural ablation of Eq.~\ref{eq:composer}).


\begin{table}[t]
\caption{Mechanism variants.}
\label{tab:mechanism}
\centering
\scriptsize
\begin{tabular}{@{}lrrrr@{}}
\toprule
\textbf{Variant} & \textbf{OFR} & \textbf{PIR} & \textbf{PIR-h} & \textbf{EFS} \\
\midrule
Stable evidence-chain & 0.5151 & 0.4792 & 0.2684 & 0.3475 \\
Fresh owner mining & 0.5991 & 0.4659 & 0.2308 & 0.3648 \\
\textbf{\paltrace{}} & \textbf{0.6057} & 0.4792 & 0.2684 & \textbf{0.3764} \\
\bottomrule
\end{tabular}
\vspace{-10px}
\end{table}

\noindent\textbf{Finding 5.} \emph{The mechanism evidence supports the state-separation hypothesis.} Removing only the composer's discard step---i.e., letting the owner-mining stage emit person rows alongside facts (\emph{Fresh owner mining})---reduces PIR from $0.4792$ to $0.4659$ and PIR-hard from $0.2684$ to $0.2308$ with the same backbone, prompts, and evidence as \paltrace{}. The drift is consistent with a predictable side effect of generating identity-shaped tokens under fresh prompt context. \paltrace{} routes the same owner-mining output through Eq.~\ref{eq:composer} and recovers the stable identity rows verbatim, while OFR rises $+9.1$ points and EFS $+2.9$ points over the \emph{Stable evidence-chain} envelope. Two architectural roles fall out cleanly: type-open mining produces the OFR/EFS gain; the composer prevents that gain from costing identity. This supports Prediction P1's mechanism-level form---the gains are state-disciplined, not call-count-disciplined---and supplies the empirical signature of Eq.~\ref{eq:composer} promised in Section~\ref{subsec:paltrace}.

\vspace{-5px}
\subsection{RQ6: How Robust and Costly is the Reference Design?}
\label{subsec:rq6_robust}

Table~\ref{tab:backbones} reports the full metric suite across every completed full-50 backbone run, with GPT-5.4 fixed in advance as the primary reporting backbone. Eight of nine backbones exceed every non-\paltrace{} baseline of Table~\ref{tab:main_results} on OFR, PIR, and EFS, supporting the view that the framework contributes beyond the choice of LLM backbone. The frontier-class Gemini~3.1~Pro Preview backbone further extends \paltrace{} to OFR 0.6142, PIR 0.5648, PIR-hard 0.3790, and EFS 0.4206---widening the gap to baselines and showing that the headline GPT-5.4 numbers, while strong, are not the peak observed configuration.

\begin{table*}[t]
\caption{\paltrace{} backbone robustness across completed full-50 runs; Qwen3.6 is informational because it shares the judge family.}
\vspace{-5px}
\label{tab:backbones}
\centering
\scriptsize
\begin{tabular}{@{}lrrrrrrrr@{}}
\toprule
\textbf{Backbone} & \textbf{OFR} & \textbf{PIR} & \textbf{PIR-h} & \textbf{PRR-ID} & \textbf{EFS} & \textbf{ECE} & \textbf{\nllm{}} & \textbf{$\Delta$EFS} \\
\midrule
GPT-5.4 (primary) & 0.6057 & 0.4792 & 0.2684 & 0.8384 & 0.3764 & 0.3236 & 7.4400 & 0.0000 \\
Qwen3.6-35B-A3B & 0.5381 & 0.4373 & 0.2245 & 0.8079 & 0.3188 & 0.3691 & 7.5000 & -0.0576 \\
Claude Sonnet 4.5 & 0.5991 & 0.4498 & 0.2219 & 0.8529 & 0.3551 & 0.3352 & 7.4400 & -0.0213 \\
GPT-5.4 mini & 0.5571 & 0.4239 & 0.2276 & 0.7973 & 0.3222 & 0.3793 & 7.4200 & -0.0542 \\
DeepSeek V4 Flash & 0.5848 & 0.4836 & 0.2409 & 0.8329 & 0.3620 & 0.3290 & 7.3400 & -0.0144 \\
Gemma 4 26B-A4B IT & 0.5014 & 0.3962 & 0.2444 & 0.7719 & 0.2867 & 0.3841 & 7.3600 & -0.0897 \\
Nemotron 3 Nano FP8 & 0.4590 & 0.4023 & 0.1837 & 0.6082 & 0.2341 & 0.3826 & 7.5400 & -0.1423 \\
Gemini 3.1 Flash-Lite & 0.5967 & 0.4367 & 0.2179 & 0.7888 & 0.3330 & 0.3641 & 7.4200 & -0.0434 \\
Gemini 3.1 Pro Preview & 0.6142 & 0.5648 & 0.3790 & 0.8851 & 0.4206 & 0.2910 & 7.4800 & +0.0442 \\
\bottomrule
\end{tabular}
\vspace{-11px}
\end{table*}

We also test whether the official semantic judge drives the ordering. On a stratified 10-user subset scored by an independent GPT-OSS-120B judge, method-rank correlations with the official judge are Spearman 0.976/1.000/1.000/0.905 for OFR/PIR/PIR-hard/EFS, and the \paltrace{} versus compute-scaled long-context deltas keep direction under both judges---making the main ordering unlikely to be a single-judge artifact.

\noindent\textbf{Finding 6.} \paltrace{} is not tied to a single model, but the quality-cost frontier is real (Table~\ref{tab:resources}). The primary run uses 7.44 calls/user, 531K tokens/user, and 138.5 sec/user---higher than the single-call long-context and RAG baselines (~30 sec/user) but with substantially higher EFS and PIR. The two compute-matched diagnostics use comparable budgets yet still trail \paltrace{} by 17.5 EFS points and 40.4 PIR points. Judge calls average 47.14/user for \paltrace{} because matched claims and cited evidence are evaluated semantically.

\begin{table}[t]
\caption{Resource diagnostics; \paltrace{} token/runtime values use the dedicated full-50 token-audit run.}
\label{tab:resources}
\centering
\scriptsize
\begin{tabular}{@{}lrrrr@{}}
\toprule
\textbf{System} & \textbf{\nllm{}} & \textbf{Sec/user} & \textbf{Tok/user} & \textbf{Parse fail} \\
\midrule
\paltrace{} & 7.44 & 138.50 & 531,443.6 & 0.0000 \\
Long-context & 1.00 & 30.51 & 43,456.8 & 0.0000 \\
Compute-scaled LC & 7.04 & 479.21 & 458,980.4 & 0.0400 \\
Multimodal RAG & 1.00 & 32.06 & 67,124.6 & 0.0000 \\
Generic tool-use & 2.10 & 46.05 & 74,155.1 & 0.0476 \\
Generic tool-use 7-call & 7.04 & 226.80 & 270,631.8 & 0.0400 \\
\bottomrule
\end{tabular}
\end{table}

\subsection{Error Analysis}
\label{subsec:error}

We audit 100 target-level errors: 50 from \paltrace{} and 50 from the compute-scaled long-context diagnostic. Table~\ref{tab:failures} shows that identity binding dominates both systems, but the residual patterns differ: \paltrace{} more often fragments aliases or overcommits on weakly supported social links, whereas compute-scaled long-context more often fails at the underlying face-name binding. Calibration flags are consistent with this picture: 20/50 \paltrace{} errors and 7/50 compute-scaled long-context errors are high-confidence wrong or unsupported, while neither sample omits citations. The remaining challenges are therefore alias consolidation, relation disambiguation, evidence selection, and persistent identity state.


\section{Related Work and Positioning}
\label{sec:related}


\noindent\textbf{Multimodal QA and long-context reasoning.} VQA~\cite{antol2015vqa,johnson2017clevr,hudson2019gqa,marino2019okvqa}, document-image~\cite{singh2019textvqa,mathew2021docvqa,biten2019stvqa,masry2022chartqa,mathew2022infographicvqa}, and long-video/egocentric/multi-image benchmarks~\cite{lei2018tvqa,xiao2021nextqa,grauman2022ego4d,damen2018epickitchens,wu2024longvideobench,fu2025videomme,song2024milebench,liu2024mibench,wang2024muirbench} evaluate prompted image, OCR/layout, or temporal comprehension. \pal{} differs by removing target-specific questions and requiring album-scale structured profiles with evidence citations.

\smallskip\noindent\textbf{Lifelog and personal media.} Lifelog campaigns~\cite{gurrin2016ntcir,dangnguyen2017imagecleflifelog,chen2025ntcir18lifelog} and personal-memory QA datasets~\cite{jiang2017visualmemoryqa,jiang2017memexqa,doherty2024memoriqa,tonellotto2023timelineqa,openlifelogqa2025} study retrieval/QA over personal visual streams, often with real data. They do not provide open-world structured owner/social reconstruction with complete face-name-relation ground truth and cited public evidence; \pal{} is complementary because synthetic construction can expose that otherwise unreleasable evaluator state.

\smallskip\noindent\textbf{Data integration and entity resolution.} Data integration~\cite{doan2012dataintegration,halevy2006data,rahm2001survey}, entity resolution~\cite{christophides2021entityresolution,elmagarmid2007duplicate,papadakis2018jedai,mudgal2018deepmatcher,li2020ditto,govind2019magellan}, data discovery/curation~\cite{stonebraker2013datacuration,deng2017datacivilizer,fernandez2018aurum,nargesian2018tableunion,subramaniam2025blend}, and user linkage~\cite{shu2017user,farnadi2018user} largely assume structured attributes or social metadata. \pal{} extends this lineage to \emph{perceptual entity resolution}: joins must be created from face clusters, OCR strings, captions, timestamps, locations, and event co-occurrence, so identity discovery, binding, and relation recovery are evaluated separately.

\begin{table}[t]
\caption{Failure-mode incidence in 100 sampled errors; P/CL denotes \paltrace{}/compute-scaled long-context counts.}
\label{tab:failures}
\centering
\scriptsize
\setlength{\tabcolsep}{2pt}
\begin{tabular}{@{}p{0.21\linewidth}p{0.34\linewidth}p{0.20\linewidth}p{0.13\linewidth}@{}}
\toprule
\textbf{Failure mode} & \textbf{Typical symptom} & \textbf{Metrics affected} & \shortstack{\textbf{Inc.}\\\textbf{P/CL}} \\
\midrule
Identity binding & Face/name absent or attached to the wrong public identity & PIR, PIR-hard & \shortstack{31/50\\39/50} \\
Alias fragmentation & Nickname and full name become separate people & PIR, PRR-ID & \shortstack{9/50\\0/50} \\
Missed evidence & Recoverable sparse evidence is not promoted into output & OFR, PIR & \shortstack{4/50\\8/50} \\
Temporal specificity loss & Month, recurrence, group, or location qualifier is lost & OFR, EFS & \shortstack{3/50\\3/50} \\
Relation overreach & Relation inferred from repeated co-presence alone & PRR-ID, EFS & \shortstack{2/50\\0/50} \\
Evidence mismatch & Correct claim cites adjacent but insufficient photos & EFS & \shortstack{1/50\\0/50} \\
\bottomrule
\end{tabular}
\vspace{-11px}
\end{table}

\smallskip\noindent\textbf{Benchmark methodology and agentic systems.} Benchmark methodology emphasizes documentation, reproducibility, and holistic evaluation~\cite{bender2018datastatements,gebru2021datasheets,holland2018datanutrition,mitchell2019modelcards,srivastava2023bigbench,liang2023helm,kiela2021dynabench,mazumder2023dataperf,koh2021wilds}. \pal{} adds a construction-time evidence contract for synthetic multimodal data: private targets are compiled into public observations and audited through separated views. Agentic, memory, and RAG systems~\cite{guu2020realm,lewis2020rag,gao2023retrievalaugmentedmultimodal,nakano2021webgpt,yao2022webshop,yao2023react,park2023generativeagents,packer2023memgpt,lyu2026photocraft,schick2023toolformer,asai2024selfrag,shinn2023reflexion,sun2024benchmark,wu2024autogen,wang2024survey} supply baselines; \pal{} maps their failures with structured outputs, identity-conditioned metrics, and evidence citations.


\section{Discussion and Responsible Use}
\label{sec:discussion}

\subsection{What \pal{} Measures and What Transfers}
\label{subsec:what_measures}
\label{subsec:why_synthetic}

\pal{} measures data-engineering capabilities over multimodal personal records: sparse retrieval, perceptual entity resolution, temporal integration, and provenance-aware structured prediction. It is not a deployable profiling system and is limited to public observations within a synthetic release. Releasing real albums with complete owner profiles, social graphs, face-name mappings, and per-target evidence would violate privacy: the audit found a median of only 21\% public-shareable photos and 94/120 fact chains relying on sensitive evidence. Synthetic construction is therefore the mechanism that makes the benchmark contract possible: it gives the evaluator complete private ground truth while exposing systems only to audited public records. The separation audit and the caption-swap probe reduce shortcut concerns, but they do not eliminate them.

The benchmark should therefore be read as a controlled reconstruction test under a public-record contract, not as evidence that personal profiling should be deployed on real albums. Its diagnostic value is to separate claim discovery, identity binding, and evidence support: high OFR without PIR indicates owner summarization without social reconstruction; high PRR-ID with low PIR indicates relation labeling after the hard binding step; low EFS indicates that a plausible claim is not yet auditable.

\smallskip\noindent\textbf{Scope of transfer.} We expect qualitative findings to transfer more than absolute scores: identity binding remains harder than relation labeling once identity is correct, face evidence remains structurally necessary, and hard social binding remains the main bottleneck. The Realism Stress Suite supports this ordering under joint caption, OCR, location, and face degradation. Real albums may shift absolute operating points through sparser text or noisier clustering. We do not claim that absolute metric values, social-neighborhood-dependent identity scores, or costs carry over unchanged to real albums. The public-record contract also excludes web, address-book, or demographic enrichment that would guess beyond album evidence.

\subsection{Limitations and Responsible Release}
\label{subsec:limitations}

Several limitations remain. \pal{} does not exhaustively model all real-album messiness; its OCR density is higher than in the audit (82.4\% vs.\ 17\%), so absolute OFR/EFS on real albums should not be expected to match. The public/private audit rules out identifier- and field-level leakage but cannot exclude every semantic shortcut, and 50 users are sufficient for benchmarking but not for rare demographics or long-tail cultural variation. The official metrics are target-centered and judge-assisted; they are designed to compare systems under a fixed contract, not to certify every free-form extra claim a system may emit. Regenerating the corpus with weaker image models than Nano Banana 2 may reduce evidence-realization rates, and hosted image-generation APIs may change over time; the release therefore freezes rendered images, public measurements, manifests, and evaluation targets, while documenting generation prompts and scripts for from-scratch regeneration. All users, photos, identities, and profiles are synthetic. The private evaluator view is evaluation-only; tuning on hidden targets, ID maps, or evidence paths would invalidate the contract. The supplemental package releases generation scripts, manifests, evaluation code, cached judge outputs, and experiment manifests to support reproducibility within this synthetic-data boundary.

\subsection{Conclusion}
\label{subsec:conclusion}

We presented \pal{}, a benchmark for evidence-grounded profile reconstruction from longitudinal personal albums with auditable public/private separation. The benchmark frames personal albums as weak-schema multimodal databases and exposes a joint data-management problem spanning sparse retrieval, perceptual entity resolution, temporal reasoning, and faithful evidence citation. Its public/private contract makes otherwise unreleasable ground truth evaluable, while keeping system inputs restricted to observable records. Across baselines and diagnostics, \paltrace{} suggests that state-disciplined composition matters more than raw call count, while hard identity resolution remains open.

\clearpage
\section*{AI-Generated Content Acknowledgement}
The authors used AI-based tools for writing polish and grammar editing. Nano Banana 2 was also used as part of the benchmark construction pipeline described in Section~\ref{sec:compiler} to render synthetic album images; this construction-time use is documented in the released artifact. Public benchmark fields are derived from perception measurements rather than copied from generation prompts. All claims, experimental results, citations, and final text were reviewed and are the responsibility of the authors.

\balance
\bibliographystyle{IEEEtran}
\bibliography{refs}

\end{document}